\documentclass[sigconf]{acmart}
\newcommand{\method}{\textit{GraphPrompter}\xspace}

\newcommand{\best}[1]{\textbf{#1}}
\newcommand{\second}[1]{\underline{#1}}
\usepackage{colortbl}
\usepackage{amsmath}
\usepackage[title]{appendix}
\usepackage{xcolor}
\usepackage{hyperref}       
\usepackage{url}            
\usepackage{booktabs}       
\usepackage{amsfonts}       
\usepackage{nicefrac}       
\usepackage{microtype}      
\usepackage{graphicx}
\usepackage{amsmath}
\usepackage{graphicx}
\usepackage{caption}
\usepackage{subcaption}
\usepackage{nicefrac}       
\usepackage{microtype}      
\usepackage{graphicx}
\usepackage{layouts}
\usepackage{caption}
\usepackage{enumitem}

\usepackage[ruled,vlined,linesnumbered]{algorithm2e}
\usepackage[detect-none]{siunitx}
\AtBeginDocument{%
  \providecommand\BibTeX{{%
    \normalfont B\kern-0.5em{\scshape i\kern-0.25em b}\kern-0.8em\TeX}}}

\copyrightyear{2024}
\acmYear{2024}
\setcopyright{acmlicensed}
\acmConference[WWW '24 Companion] {Companion Proceedings of the ACM Web Conference 2024}{May 13--17, 2024}{Singapore, Singapore.}
\acmBooktitle{Companion Proceedings of the ACM Web Conference 2024 (WWW '24 Companion), May 13--17, 2024, Singapore, Singapore}
\acmISBN{979-8-4007-0172-6/24/05}
\acmDOI{10.1145/3589335.3651476}

\begin{document}

\title{Can we Soft Prompt LLMs for Graph Learning Tasks?}

\author{Zheyuan Liu}
\authornote{Equally contributed.}
\affiliation{
  \institution{University of Notre Dame}
    \country{}
}
\email{zliu29@nd.edu}

\author{Xiaoxin He}
\authornotemark[1]
\affiliation{
  \institution{National University of Singapore}
  \country{}
} 
\email{he.xiaoxin@u.nus.edu}

\author{Yijun Tian}
\authornotemark[1]
 \affiliation{
  \institution{University of Notre Dame}
\country{}
}
\email{yijun.tian@nd.edu}

\author{Nitesh V. Chawla}
 \affiliation{
  \institution{University of Notre Dame}
\country{}
}
\email{nchawla@nd.edu}




\begin{abstract}
Graph plays an important role in representing complex relationships in real-world applications such as social networks, biological data and citation networks. In recent years, Large Language Models (LLMs) have achieved tremendous success in various domains, which makes applying LLMs to graphs particularly appealing.
However, directly applying LLMs to graph modalities presents unique challenges due to the discrepancy and mismatch between the graph and text modalities.
Hence, to further investigate LLMs' potential for comprehending graph information, we introduce \textbf{\method}, a novel framework designed to align graph information with LLMs via soft prompts. Specifically, \method consists of two main components: a graph neural network to encode complex graph information and an LLM that effectively processes textual information.
Comprehensive
experiments on various benchmark datasets under node classification and link prediction tasks demonstrate the effectiveness of our proposed method. The \method framework unveils the substantial capabilities of LLMs as predictors in graph-related tasks, enabling researchers to utilize LLMs across a spectrum of real-world graph scenarios more effectively\footnote{Code is available at \url{https://github.com/franciscoliu/graphprompter}.}.

\end{abstract}




\maketitle

\section{Introduction}
Large Language Models (LLMs) have demonstrated significant success across various domains ~\cite{ chowdhery2023palm, touvron2023llama, ouyang2022training, lewkowycz2022solving, wei2024llmrec, wang2023knowledge}, largely attributed to their extensive knowledge memorized during the pretraining phase and their exceptional ability to generalize during the fine-tuning process on diverse textual datasets ~\cite{hoffmann2022training, liang2022holistic, tian2024tinyllm, carlini2022quantifying, tan2024democratizing}. This success has spurred interest in combining graph neural networks (GNNs) with LLMs to enhance their capabilities in understanding and modeling graphs~\cite{yang2021graphformers, zhang2022metadata, ostendorff2022neighborhood, tian2024gnp, he2024g}, including implementing LLMs as encoders to process features within GNNs \cite{chai2023graphllm, mavromatis2023train, he2023explanations, yu2023empower}, and employing LLMs as aligners with GNNs to enhance performance \cite{zhao2022learning, wen2023augmenting, brannon2023congrat}. 

However, directly applying LLMs to graph modalities presents unique challenges due to the discrepancy and mismatch between the graph and text modalities. For example, existing works primarily map the graph into text, ignoring the irrelevant information and noises pertinent to the graphs. This oversight results in an inadequate understanding of crucial structural knowledge within the graphs. Furthermore, the challenge is amplified when dealing with large graphs containing thousands or millions of nodes and edges, as this complexity hinders LLMs' ability to grasp intricate structural information.

These limitations motivate our investigation into using a graph as a soft prompt encoded by GNNs for LLMs in graph learning tasks. The intuition behind this approach is that  GNN are more proficient in aggregating and transforming information from the neighbourhood information, which could provide topological information to LLMs. Additionally, it can guide the LLM in selecting relevant information from textual input and control the generation process for token generation. Specifically, in this work, we aim to examine the efficacy of a soft graph prompt in informing the LLM's predictions. Hence, a natural yet pivotal research question arise: \textit{Can we soft prompt LLMs for graph leanring tasks?}

\begin{figure}[t]
    \centering
    \includegraphics[scale=0.5]{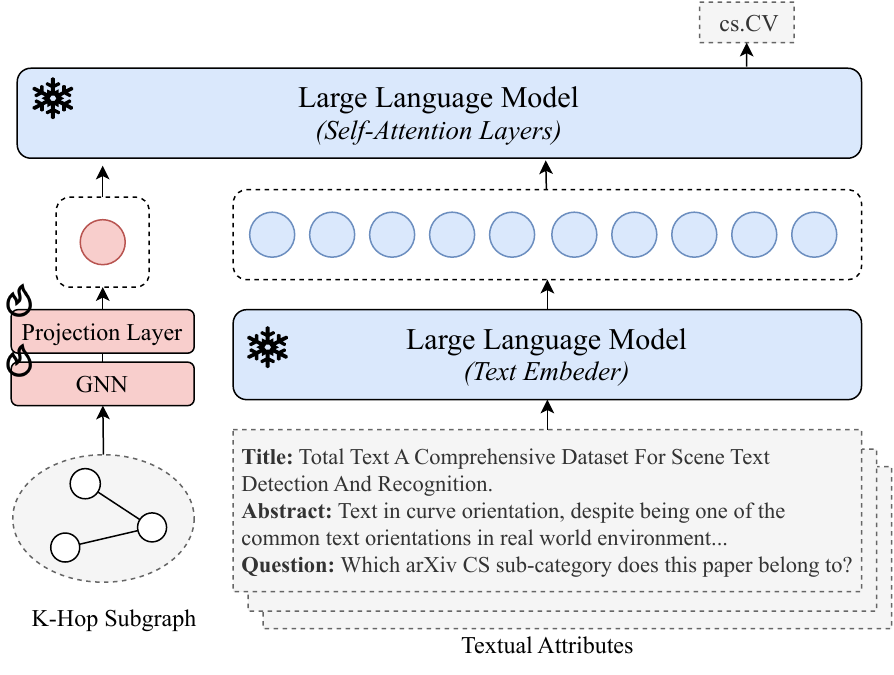}
    \caption{Illustration of the proposed \method for node classification task. The process involves extracting a k-hop subgraph for each node, feeding it into a GNN followed by a projection layer. Simultaneously, the textual attributes associated with each node are processed by the text embedder. The resulting node embedding is then concatenated with text embeddings, serving as a soft prompt to guide the LLM for graph learning tasks.}
    \vspace{-0.3in}
    \label{fig: overview}
\end{figure}

To answer this question, we introduce {\method}, a novel framework that combines the strength of GNNs and LLMs to process and comprehend graph-structured data. In particular, \method capitalizes on the frozen LLM as a robust feature extractor that taps into its vast pretraining knowledge, thus enabling us to avoid extensive task-specific fine-tuning. In parallel, the GNN works on the graph to produce node embeddings, which are later concatenated with a prompt instruction to guide LLMs for graph learning tasks. The LLM, due to its powerful autoregressive nature, generates a language response based on the fused graph and text information, effectively turning the LLM into a powerful tool for graph understanding tasks. See Figure~\ref{fig: overview} for an illustration.

This hybrid approach is particularly suitable for graphs with textual attributes (i.e., textual graphs), which requires an understanding of both the textual content and the graph structure, such as identifying the subcategory of an academic paper based on its citation network and abstract.  
We experimentally demonstrate that our framework is capable of prompting LLMs for graph learning tasks on five benchmark datasets under node classification and link prediction tasks. It showcases the potential for significant advancements in the use of LLMs for complex data structures beyond traditional text, opening up new avenues for research and application in the realm of AI assistants capable of intricate graph comprehension. Our main contributions are as follows:
\begin{itemize}
    \item To the best of our knowledge, this is the very first work investigating whether LLMs can understand graph learning tasks via soft prompting.
    \item We propose \method, a novel plug-and-play framework that first employ GNN to get node representations from the textual graph. Then the obtained embeddings are concatenated with a prompt instruction to guide LLMs for graph learning tasks. 
    \item Extensive experiments demonstrate the effectiveness of our
    proposed framework under both node classification and link prediction tasks across various graph benchmarks.
\end{itemize}



\vspace{-0.15in}
\section{Method}

To enhance the alignment of graph knowledge with LLM, we introduce \method, a plug-and-play pipeline that fuses post-processed node embeddings with LLMs. This integration is designed to leverage the rich semantic context LLMs provide, enhancing the interpretability and utility of graph representations, as illustrated in Figure ~\ref{fig: overview}.

\subsection{Graph Section}

A graph can be represented as $G = (V, E)$, where $V$ is a set of vertices, and $E$ is a set of edges. For each node $v_i \in V$, we define the 3-hop subgraph $G_{s_i}$ as the induced subgraph containing all nodes within three hops from $v_i$. Subsequently, a GNN is employed to compute the node embeddings $X_i$ for each of these subgraphs $G_{s_i}$, as follows:
\begin{equation}
    X_i = \text{GNN} (G_{s_i}) \in \mathbb{R}^{d_g},
\end{equation}
where $d_g$ is the hidden dimension of the GNN. Each embedding $X_i$ captures the structural information of the corresponding subgraph, providing a rich representation of the topological features associated with $v_i$. 

We employ a projection layer, specifically, a multilayer perceptron (MLP), to align the node embedding with the vector space of the LLM:
\begin{equation}
    \hat X_i = \text{MLP}(X_i) \in \mathbb{R}^{d_l},
\end{equation}
where $d_l$ is the hidden dimension of the LLM.

Both the GNN and the projection layer are trained in an end-to-end manner, ensuring that the embeddings $X_i$ are discriminative for the downstream task. The rationale behind this approach is twofold: first, to leverage the GNN's inherent capacity to aggregate and transform local neighborhood information into meaningful embeddings; and second, to enable the subsequent integration with the LLM, which interprets these embeddings as a soft prompt in conjunction with textual data. 

\subsection{LLM Section}

After encoding the graph structure, the \method proceeds to process the textual information associated with each node. This is where the power of the LLM comes into play, as it is kept frozen for efficient fine-tuning while preserving the LLM's pre-trained knowledge. For each node $v_i$, given its associated textual attribute $T_i$ (e.g., title and abstract of a paper in a citation graph),  we tokenize $T_i$ using the frozen tokenizer of the LLM as follows:
\begin{align}
    T_\text{tokens} = \text{Tokenizer}(T_i).
\end{align}
The tokenizer converts the text into a sequence of discrete tokens $T_\text{tokens}$ in the LLM's vocabulary. Subsequently, these tokens are then embedded into a continuous space:
\begin{align}
    T_\text{emb} = \text{Embed}(T_\text{tokens}) \in \mathbb{R}^{M \times d_l},
\end{align}
where $M$ is the sequence length and $d_l$ is the hidden dimension of the LLM. These embeddings are designed to capture the semantic meaning of the text, complementing the structural information provided by the GNN. 

Next, we concatenate the node embeddings and the text embeddings, denoted as $[\hat X_i, T_\text{emb}]$, which will go through the self-attention layers in the LLM as usual.

The motivation for this step is to ensure that the LLM can process the rich semantic content of the text alongside the graph embeddings. By encoding the text attributes into a compatible format, we aim to capitalize on the LLM's advanced understanding of natural language, which is crucial for tasks that require a combination of structural and textual data interpretation, such as node classification in citation networks. Here, we point out the key distinction between node classification and link prediction: node classification evaluates information from a single node, whereas link prediction considers the attributes of two nodes, specifically their origin and destination.

\section{Experiments}

\begin{figure}
\centering
\begin{subfigure}{0.49\linewidth}
    \centering
    \includegraphics[width=\linewidth]{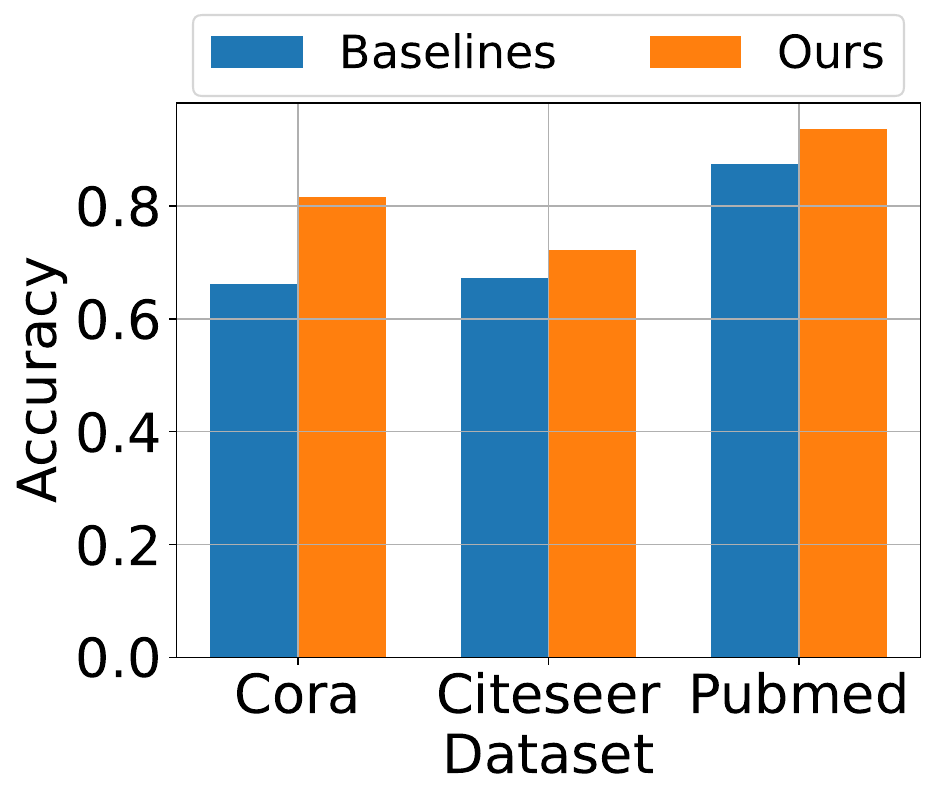}
    \subcaption{Dense (Title + Abstract)}
    \label{fig:dense}
\end{subfigure}
\begin{subfigure}{0.49\linewidth}
    \centering
    \includegraphics[width=\linewidth]{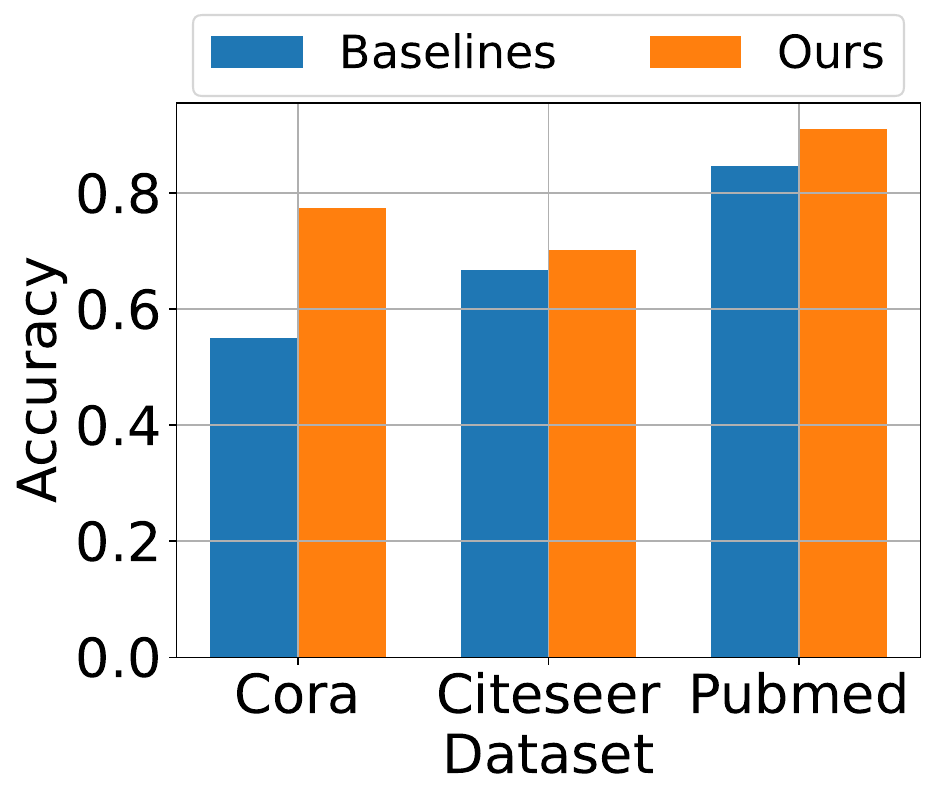}
    \subcaption{Sparse (Title Only)}
    \label{fig:sparse}
 \end{subfigure}
\vspace{-0.1in}
\caption{Node classification experiments comparing the proposed \method with baseline methods (i.e., soft prompt tuning and fine-tuning). The $x$ axis shows dataset types, and the $y$ axis displays accuracy scores. Figure \ref{fig:dense} illustrates a dense semantic setting including both the title and abstract of a paper within the node embeddings, while Figure \ref{fig:sparse} illustrates  a sparse semantic setting with title only.}
\vspace{-0.25in}
\label{fig:dense_sparse}
\end{figure}

In this section, we conduct extensive experiments to validate the effectiveness of the \method. Specifically, our experiments aim to address the central question: \textbf{Can we soft prompt LLMs for graph learning tasks?}
\vspace{-0.1in}
\subsection{Experiment setups}
\subsubsection{\textbf{Datasets and baseline models.}} Our experiments mainly focus on node classification and link prediction using various graph benchmarks: Cora, Citeseer, Pubmed, Ogbn-arxiv, and Ogbn-products. For Ogbn-products, given its substantial scale of 2 million nodes and 61 million edges, we have employed a node sampling strategy to obtain a subgraph containing 54k nodes and 74k edges. 
We compare our approach with three kinds of baseline models: 1) Pure GNN; 2) Frozen LLM, which includes zero shot and soft prompt tuning; and 3) Tuned LLM with LoRA. We employ the LLAMA2-7B~\cite{touvron2023llama} as the LLM backbone. For the pure GNN method, we use GAT~\cite{velivckovic2017graph}. In the LLM-frozen setting, for zero-shot, we use a frozen LLM for direct node classification/edge prediction using textual attributes of the nodes. In soft prompt tuning, we keep the parameters of the LLM frozen and adjust only the prompt. In LLM-tuned, the original LLM parameters are updated for downstream tasks by utilizing LoRA.

\vspace{-0.1in}
\subsubsection{\textbf{Implementation Details.}} We report the mean of five independent runs with different seeds. The experiments are conducted on 2 A100 GPUs (80GB). 

\section{Experiments Result}

To answer the central question, we conduct experiments across various graph benchmark to comprehensively evaluate the effectiveness of \method. The performance is reported in Table ~\ref{tab: node_class_result} and Table ~\ref{tab: link_pred_result}.

\subsection{Node Classification Task}
\begin{table}[t]
\resizebox{1.0\linewidth}{!}
{
\begin{tabular}{l|ccccc}
\toprule[1pt]
Method & Cora & Citeseer & Pubmed & Ogbn-arxiv  & Ogbn-products \\
\cline{1-6}
\midrule
\multicolumn{6}{c}{Node Classification Accuracy (\%) $\textcolor{blue}{\uparrow}$} \\
\midrule
GAT& \best{84.69} & 70.78 & 84.09 & 71.82 & 70.52\\
Zero-Shot  &43.31 & 29.22 & 91.39 & 44.23 & 15.05 \\
Soft Prompt Tuning & 70.31 & 70.97 & 91.45 & 71.99& 75.14 \\ 
Fine-tuning + LoRA & 75.97 & \second{73.45} & \second{94.68} & 74.58 & 78.99\\
\rowcolor{gray!12}Subgraph Prompt Tuning& 80.17  &72.29 & 93.84 & \second{75.04} & 75.30\\
\rowcolor{gray!12}\method + LoRA & \second{80.26} & \best{73.61} & \best{94.80} & \best{75.61} & \best{79.54}\\
\midrule
\bottomrule[1pt]
\end{tabular}
}
\caption{Node classification result of our proposed \method, with a number of baselines under various graph benchmarks: Cora, Citeseer, Pubmed, Ogbn-arxiv and Ogbn-products. For each model, we present the mean accuracy from five independent runs, with both the title and abstract of papers provided as input along with specific instructions. \best{Bold} indicates the best performance and \second{underline} indicates the runner-up.}
\label{tab: node_class_result}
\vspace{-0.4in}
\end{table}

In Table \ref{tab: node_class_result}, we present the results of node classification experiments conducted on a variety of benchmark datasets, where both title and abstract information are provided to LLM alongside detailed instructions. As it can be seen from the table, the integration of the proposed method (\method) with LoRA consistently surpasses the performance of other baseline methods across diverse benchmarks. In particular, \method with LoRA in the PubMed and Citeseer datasets with top accuracies of 94.80 \% and 73.61 \%, respectively. The naive fine-tuning with LoRA approach typically appears as the runner-up and has small performance gap with \method. Though subgraph prompt tuning approach has relatively close accuracy score with naive fine-tuning, it still not as effective as fine-tuning on node classification task. Furthermore, we observe that zero-shot generally ranks lowest in performance across all benchmark datasets, with the exception of PubMed dataset. This reflects the limitation of LLMs in capturing graph knowledge without the auxiliary processing provided by GNNs. Additionally, two distinct experimental setups are illustrated in Figure \ref{fig:dense_sparse}, where Figure \ref{fig:dense} encompasses both title and abstract within the node embeddings, and Figure \ref{fig:sparse} restricts the node embeddings to only the title. The comparative analysis illustrated in the figures further illustrates the superior average performance of the proposed method (\method), which combines subgraph prompt tuning and \method, in comparison to other baseline approaches, specifically soft prompt tuning and fine-tuning methodologies. 

\subsection{Link Prediction Task}
In Table \ref{tab: link_pred_result}, we present the results of link prediction experiments conducted on a variety of benchmark datasets, where only the title is provided to LLM. As seen in Table ~\ref{tab: link_pred_result}, subgraph prompt tuning approach consistently surpasses baseline models in link prediction tasks on various benchmarks. It shows remarkable performance especially on the Citeseer dataset, achieving a 93.49 \% accuracy rate. \method + LoRA is usually the runner-up model for the majority case. Notably, similar to node classification task, the Zero-Shot approach generally scores lowest, underscoring the necessity of GNN integration for LLMs to effectively process and understand graph data. It is notable that under both node classification and link prediction, \method can continue outperforming traditional GNN and soft prompting techniques. This observation indicates that \method, is particularly adept at capturing and utilizing the complex intersection between graph components and textual information, leading to a better performance on various graph related tasks.

\begin{table}[t]

\resizebox{1.0\linewidth}{!}
{
\begin{tabular}{l|ccccc}
\toprule[1pt]
Method & Cora & Citeseer & Pubmed & Ogbn-arxiv  & Ogbn-products \\
\cline{1-6}
\midrule
\multicolumn{6}{c}{Link Prediction Accuracy (\%) $\textcolor{blue}{\uparrow}$ } \\
\midrule
GAT& \best{90.71} & 87.40 & 86.18  & 72.93 & 65.67\\
Zero-Shot &  21.94 &7.69  & 10.23 & 34.41 & 7.47 \\
Soft Prompt Tuning & 86.77 & 88.87 & 83.98 & 69.13& 62.78 \\ 
Fine-tuning + LoRA & 87.58  & 87.91& 81.33 & 70.10& 67.31\\
\rowcolor{gray!12}Subgraph Prompt Tuning& 89.15  & \best{93.49} & \best{87.20} & \best{75.28} & \second{68.18}\\
\rowcolor{gray!12}\method + LoRA & \second{90.10} & \second{91.67} & \second{86.49} & \second{73.21} & \best{69.55}\\
\midrule
\bottomrule[1pt]
\end{tabular}
}
\vspace{0.1in}
\caption{Link prediction result of our proposed \method, with a number of baselines under various graph benchmarks: Cora, Citeseer, Pubmed, Ogbn-arxiv and Ogbn-products. For each model, we present the mean accuracy derived from five independent runs conducted on the link prediction task, with only the abstract of papers are provided as input. \best{Bold} indicates the best performance and \second{underline} indicates the runner-up.}
\vspace{-0.4in}
\label{tab: link_pred_result}
\end{table}

\section{Analysis}
The experiments on benchmark datasets for node classification and link prediction tasks demonstrate LLMs' adaptability and efficiency in graph learning, enabled by our proposed soft graph prompting strategies. Here, we summarize the key findings:
\begin{itemize}
    \item \textbf{Performance Benchmark}: Overall, our proposed method ranks either as the best or runner-up across all benchmark datasets for both tasks. This consistently high performance affirms a positive response to our central question: we can effectively soft prompt for LLMs in graph learning tasks.
    \item \textbf{Methodological Superiority}: The fine-tuning approach, while competitive, displayed a marginal performance gap when compared to \method. This indicates that while traditional fine-tuning remains effective, the specialized soft graph prompt tuning strategy designed for graph data offers a more potent solution for enhancing LLM's graph learning capabilities.
    \item \textbf{Challenges with Zero-Shot Learning}: 
    The zero-shot approach consistently ranked the lowest across all benchmarks for both tasks. This reflects the challenges LLMs face in comprehending graph structure and semantics, highlighting the necessity for tailored approaches like \method in graph learning tasks.
\end{itemize}

\vspace{-0.1in}
\section{Conclusion}
In conclusion, our study reveals the significant potential of leveraging LLMs for interpreting graph structures through a prompt tuning methodology. To facilitate this study, we presents \method, a novel plug-and-play framework that integrates the capabilities of Large Language Models with the structural insights of Graph Neural Networks for the task of node classification and link prediction in graph data. Specifically, \method consists of two sections: the graph section utilizes GNN for structural insights, while the text section applies LLM to interpret node-related textual data. Extensive experiments and in-depth studies demonstrate the superiority of \method across multiple benchmark datasets.  Future work could focus on extending our pipeline to other more complex graph level tasks. 

\vspace{-0.1in}
\bibliographystyle{ACM-Reference-Format}
\bibliography{ref}

\end{document}